\documentclass{article}

\usepackage{graphics}
\usepackage[utf8]{inputenc} 
\usepackage[T1]{fontenc}    
\usepackage{url}            
\usepackage{booktabs}       
\usepackage{amsfonts}       
\usepackage{nicefrac}       
\usepackage{microtype}      
\usepackage{xcolor}         
\newcommand{\etal}{\textit{et al.}}
\usepackage{graphicx}
\usepackage{amsmath}
\usepackage{soul}
\usepackage{algorithm}
\usepackage{algpseudocode}
\usepackage[ruled,vlined,algo2e]{algorithm2e}
\usepackage{booktabs}
\usepackage{boldline}
\usepackage{siunitx}
\usepackage{wrapfig}
\usepackage{amsthm}

\newcommand{\brwonian}{\mathbf{w}}
\newcommand{\scorenet}{s_\theta}
\newcommand{\score}{\frac{\partial \log  p(x)}{\partial x} }

\usepackage{mdframed}

\newtheorem{theorem}{Theorem}

\newtheorem{proposition}{Proposition}

\DeclareMathOperator*{\argmin}{arg\,min}

\title{Score-based Generative Modeling Through Backward Stochastic Differential Equations: Inversion and Generation}

\author{%
  Zihao Wang\thanks{Corresponding: zihao.wang@ieee.org} \\
  A. A. Martinos Center for Biomedical Imaging@MGH, \\
  Harvard University.\\
  149 13th St, Charlestown, MA 02129 \\
  \texttt{zwang63@mgh.harvard.edu} \\
}

\begin{document}

\maketitle

\begin{abstract}
The proposed BSDE-based diffusion model represents a novel approach to diffusion modeling, which extends the application of stochastic differential equations (SDEs) in machine learning. Unlike traditional SDE-based diffusion models, our model can determine the initial conditions necessary to reach a desired terminal distribution by adapting an existing score function. We demonstrate the theoretical guarantees of the model, the benefits of using Lipschitz networks for score matching, and its potential applications in various areas such as diffusion inversion, conditional diffusion, and uncertainty quantification. Our work represents a contribution to the field of score-based generative learning and offers a promising direction for solving real-world problems.
\end{abstract}

\section{Introduction}
Stochastic differential equations (SDEs) have emerged as an indispensable tool for analyzing diffusion processes. The use of SDEs allows for a robust framework that can capture the inherent random fluctuations of the diffusion process, an area of interest that has been widely explored in the field of machine learning. One specific application of diffusion models, known as score matching, has gained popularity in the field of machine learning for density estimation. Density estimation is the process of estimating the underlying probability distribution of a given dataset. The effectiveness of score matching in accurately representing datasets has made it a preferred method in various domains, including image, video, audio, text, and speech synthesis, to name a few. Recent studies have demonstrated the effectiveness of score matching in various applications, such as photorealistic image synthesis \cite{saharia2022photorealistic}, zero-shot learning \cite{wang2023zeroshotlearning}, diffusion-based generative models \cite{NEURIPS2021_49ad23d1, yang2022diffusion, alcaraz2023diffusionbased}, image compression \cite{ho2022imagen}, time-series modeling \cite{tashiro2021csdi}, and more \cite{6795935,NEURIPS2020_4c5bcfec, DDPM}. SDEs are a strong tool for data modeling through diffusion models, with Song \etal \cite{song2021scorebased} proposing a framework for time-conditional score matching. De Bortol \cite{de2022riemannian} extended this to Riemannian manifolds and Huang \etal \cite{variationalSDE} demonstrated the equivalence between score matching and reverse time SDEs. Our BSDE-based diffusion framework is unique in its link to partial differential equations via the Feynman-Kac formula \cite{peng1992nonlinear}, enabling further deterministic analysis.To expand the application of diffusion models to real-world problems, we propose incorporating backward stochastic differential equations (BSDEs) into the diffusion model, which we collectively refer to as forward backward stochastic differential equations (FBSDEs). 

In contrast to standard SDE-based diffusion models \cite{song2021scorebased}, our approach assumes a pre-existing score function and a point sampled from the target terminal distribution. Our goal is to adapt the score function to attain the desired target terminal distribution using backward stochastic differential equations (BSDEs) \cite{BISMUT1973384,PARDOUX199055}. Unlike the standard SDE-based diffusion approach, our BSDE-based diffusion model allows us to obtain a deterministic solution to sample the desired terminal data point without precise statistical knowledge of it.
BSDEs originated from Bismut's work on linear control systems \cite{BISMUT1973384} and were later established to be well-posed in both linear \cite{BENSOUSSAN1983387} and nonlinear cases \cite{PARDOUX199055}. BSDEs have found widespread use in various real-world applications, including financial derivatives valuation, optimal control, risk measures, physics, and image processing \cite{Chessari_2023, BSDEFin_Kar_Pen, Hientzsch, BSDE_opti2, el1997reflected, hawkins2021value, BSDE_ris, BSDE_riskmeas, BSDE_opti1, CiCP-27-589, Bao_2019, PhysRevE.96.042123, PhysRevE.95.032418, BSDE_Applications, BSDEFin_Kar_Pen}. They are also connected to other areas \cite{Double_refl1, Double_refl2, Complex_Egs1, Complex_Egs2}. In this work, we expand the application of BSDEs in the field of machine learning by providing a theoretical foundation and empirical evidence for the diffusion model.

Our proposed framework offers both theoretical and practical contributions. Theoretically, we establish the BSDE-based diffusion framework by incorporating a Lipschitz network for learning the score function and imposing an integrability condition on the generator function. We demonstrate that incorporating spectral normalization in the score model results in better convergence and robustness. Furthermore, we establish the well-posedness of the BSDE-based diffusion model, indicating that a unique initial state and stochastic processes can be identified to attain a target using a trained Lipschitz score network.
Practically, our BSDE-based diffusion can be used to solve several real-world problems. Firstly, the BSDE can invert any arbitrary score model back into the latent representation using a given target data point, making it useful in image editing applications. Secondly, our method does not require the determination of a time point for unsupervised conditional generation, representing a significant breakthrough compared to traditional distributional perturbing techniques. Thirdly, our BSDE-based diffusion enables conditioning the generative process using the MCMC sampling method, distinct from supervised learning-based diffusion conditioning. Finally, our framework facilitates the quantification of uncertainty in a single data point, even when working with a small sample size, and provides a means for estimating marginal uncertainty in situations where the generation target is a subset of the full dataset.

\section{Theory}
\subsection{Score Matching} Our focus is on using backward stochastic differential equations (BSDEs) in probabilistic diffusion-based generative learning. In this approach, score-based models represent the distribution of a dataset through its derivative, the score function. The score function can be obtained by training a neural network on a perturbed data point obtained by randomly sampling a noise data point. The generation process involves denoising the score function into a sequence of data points with decreasing levels of noise. Song et al. unified the learning and generation processes of score matching using two SDEs: a perturbing process to estimate the score function and a reverse-time SDE to sample data points from Gaussian distributions.

\textbf{Standard SDE for Score Matching:}  Given a probability space $(\Omega, \mathbb{F}^{\brwonian}, P)$ with a Wiener process $\brwonian_s$, the stochastic equation \eqref{eq;addingnoise} describes the evolution of a data point $x_s$ over time $s \in [0,T]$, with $f(x_s,s)$ and $\sigma_s$ controlling the perturbation and noise, respectively. By training a neural network $\scorenet(x_t)$ over the time interval $[0,T]$, the score function can be learned to push an initial time state data point $x_0\sim p_0(\cdot)$ to a terminal state $x_T$ that follows a target distribution $p_T(\cdot)$.
\begin{equation}
d x_T = f(x_T, T)dt + \sigma_T d\brwonian
\label{eq;addingnoise}
\end{equation}
A neural network $\scorenet(x_t)$ can be trained to learn the score function by traversing the time interval $[0, T]$.
\begin{equation}
\mathcal{L}(\theta) = \mathbb{E}_{t \sim \mathcal{U}(0, T)} \mathbb{E}_{x_0 \sim p_0(x)} \mathbb{E}_{x_t \sim p_t(x_t|x_0)}[ \lVert\scorenet(x_t, t) - \nabla_x (\log p_t(x_t|x_0)) \rVert^2]
\label{eq;scoreloss}
\end{equation}
, where $\scorenet(x_t, t)$ learns to extract the time conditioning noise (score) $\score := -\frac{x_t - x}{\sigma_t^2}$. 

\begin{figure}[t]
    \centering
    \includegraphics[width=\textwidth]{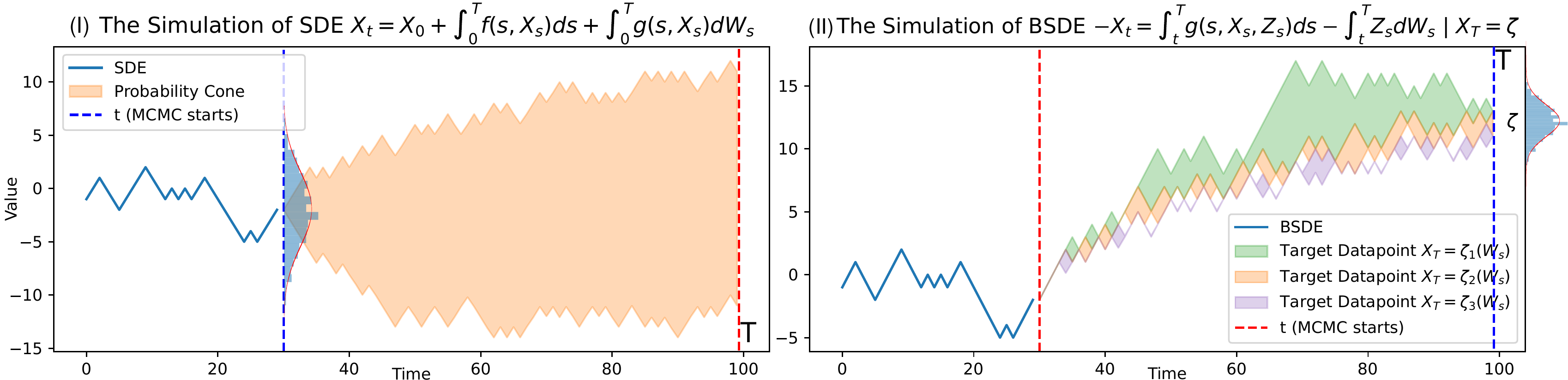}
    \caption{\textbf{Comparing SDE and BSDE Using MCMC Simulation.} Both sub-figures show the unknown state as a red vertical line and the known state as blue. The left panel displays sample paths generated by SDE, while the right panel depicts the more complex behavior of BSDE paths due to the influence of future conditions on the present state. This figure highlights the fundamental differences between the dynamics and behavior of SDE and BSDE diffusion.}
    \label{fig:comparison}
\end{figure}

\textbf{Generation with Standard Reverse Time SDE:} In the sampling process, a reverse time SDE \cite{ANDERSON1982313} $t: T \rightarrow 0$ is employed for generating a data point $x_T$ from an initial state $x_0\sim p_0(x)$:
\begin{equation}
\label{eq;reverse_derivative}
\frac{\mathrm{d} x_s}{\mathrm{d} s} = f(x_s, s) - g^2(s) \scorenet(x_s, s) + g(s) \mathrm{d} \mathbf{w}_s, \quad s \in [t,T],
\end{equation}
where $g(s) \in [0, \infty)$ controls the magnitude of the input noise, $\mathbf{w}_s$ denotes a Wiener process, and $s \in [t, T]$ denotes time starting from 0 with infinitesimal incremental to $T$. Equation \ref{eq;reverse_derivative} is used for reversing the noise for generative modeling. Figure \ref{fig:comparison}-I shows the dynamical sampling through a SDE, which is used for both mapping the data to noise distribution (SDE in Eq. \ref{eq;addingnoise}) and for reversing the noise for generative modeling (SDE in Eq. \ref{eq;reverse_derivative}).

\textbf{SDEs Solver}
To generate data points from the trained score model, we solve Eq. \ref{eq;reverse_derivative} using numerical SDE solvers such as Euler-Maruyama. However, obtaining an analytical solution for non-linear SDEs can be challenging. Several works have explored the impact of different solver methods in SDE-based diffusion, including the predictor-corrector sampler of MCMC \cite{song2021scorebased, karras2022elucidating}, Leapfrog method \cite{mao2023leapfrog}, and Heun's method \cite{jolicoeurmartineau2021gotta}.

\paragraph{BSDEs}
A Backward Stochastic Differential Equation (BSDE) is a type of SDE that models the behavior of a random process, but unlike a standard SDE, the solution is determined by working backward from a specified terminal condition. A BSDE is a backward-looking equation that describes the evolution of a stochastic process from a specified terminal condition, working backward in time to find the initial condition or an intermediate state. This is illustrated in Fig. \ref{fig:comparison}-II. In Section \ref{sec;BSDESCORE}, we will discuss how BSDEs are connected to score-based generative learning.
Let $(\Omega, \mathbb{F}^{\brwonian}, P)$ be a probability space with a Wiener process $\brwonian$. For any terminal condition $Y_T = \xi$, there exists a unique $Z$ such that a backward stochastic differential equation (BSDE) of the form $Y_t = \xi + \int_t^T \hat{f}(s, Y_s, Z_s) ds - \int_t^T Z_s d\brwonian_s$ holds.

\section{BSDEs for Score-based Generative Modeling}
\label{sec;BSDESCORE}
This work focuses on the problem of determining the initial state and adjusting the scale of diffusion to reach a known terminal state, which is the opposite scenario of typical SDEs. This is achieved through the use of BSDEs for diffusion-based generative learning, allowing for dynamic adjustment of initial conditions and stochastic paths to achieve downstream tasks. However, ensuring network continuity is crucial for convergence when using BSDEs, which is addressed in this work. The metaphor of "the sword of Damocles" is used to illustrate the challenge of network continuity.

\textbf{The Lipschitz Network for Robust Score Matching:}
Spectral normalization is a technique that rescales the parameters of a neural network to bound its Lipschitz constant \cite{miyato2018spectral}. For a neural network with layers $l=1,\dots,L$, let $W^{(l)}$ denote the weight matrix of the $l_{th}$ layer. The spectral norm of a matrix $W^{(l)}$ is defined as $|W^{(l)}|_2=\sigma_{\text{max}}(W^{(l)})$, where $\sigma_{\text{max}}(W^{(l)})$ is the largest singular value of $W^{(l)}).$ The spectral normalization constrains the spectral norm of each weight matrix by normalizing it with its largest singular value: $\tilde{W}^{(l)}=\frac{W^{(l)}}{|W^{(l)}|_2}$
With spectral normalization, we ensure that the Lipschitz constant $K$ is bounded by $\prod{l=1}^L |W^{(l)}|_2$, where $K$ is the Lipschitz constant of the neural network. The largest singular value $\sigma_{\text{max}}(W^{(l)})$ can be approximated using the power method. By applying spectral normalization to all the convolutional layers of the score network $\scorenet$, we can bound the Lipschitz constant of the network and ensure its continuity.

\subsection{BSDE-based Diffusion: The Future Creates the Present}
\paragraph{The General form of BSDE-based Diffusion}

We define the BSDE-based diffusion model as follows: Let $X_t$, $Y_t$, and $Z_t$ be three square integrable stochastic processes for $t \in [0, T]$, where $\brwonian$ is a Wiener process. Among them, $Y_t$ is a process whose terminal condition is i.i.d. to the terminal time distribution of $X_t$ at time $T$: $X_T, X_T \sim p(x)$. Additionally, $X_t$ has a score function approximated by $\scorenet$, and $\hat{f}$ is an L-Lipschitz continuous function.
\begin{equation}
Y_t = \xi + \int_t^T \hat{f}(\scorenet(Y_s, s), Z_s) dt - \int_t^T Z_s d\brwonian_s , ~~~~s \in [0, T],~t\in [0,T],~t\leq s,
\end{equation}
\label{def;bsdescore}

It is notable that the terminal condition $\displaystyle{\lim_{t\rightarrow T}} Y_t = \xi \sim p(x)$ is intrinsic to the BSDE diffusion model definition, while the stochastic processes $Y_t$ and $Z_t$ remain unspecified.

Our BSDE-based diffusion model \ref{def;bsdescore} diverges from the conventional SDE-based diffusion models \cite{song2021scorebased,songNoise} in its approach. The latter diffuses the initial condition (i.e., noise) $X_0 \sim \mathcal{N}(0,\sigma_0^2)$ to the distribution of the terminal datapoint (i.e., future) $X_T \sim p(x)$ through the gradient information $\score$ of the dataset distribution. Conversely, our BSDE-based diffusion \ref{def;bsdescore} seeks to establish an initial condition $Y_{t=0} \sim \mathcal{N}(\mu_0|\sigma_0^2)$ and a conditioning process $Z_t$ that satisfy the terminal condition $Y_T \sim p(x)$. Once we identify the two stochastic processes $(Y_t, Z_t)$ for a given datapoint $Y_T=\xi$, we can conduct several analyses on the score model and manipulate it by altering either $Y_{t=0}$, $Z_t$, or both.

\paragraph{Solution Existence and Uniqueness for BSDE-based Diffusion}
The well-posedness of the BSDE-based diffusion model defined in Eq. \ref{def;bsdescore} is guaranteed by the continuity of the generator function, as established in previous studies \cite{PARDOUX199055, pengwellposed}. Hence, we impose two conditions on the form of the function $\hat{f}(\scorenet(Y_t, t), Z_t)$, which are essential for ensuring the existence and uniqueness of the solution for the BSDE-based diffusion model.
\begin{figure}[htbp]
  \centering

        \begin{algorithm}[H]
        \label{numsolver}
        \SetAlgoLined
        \DontPrintSemicolon
        \SetKwInput{KwInput}{Input}
        \SetKwInput{KwOutput}{Output}
        \KwInput{Initial parameters \(\alpha\) and $y_n^{n,M}(\cdot) =\Phi$}
        \KwOutput{Optimal parameters \(\alpha\)}
        \KwData{Brownian motion \(\left\{\brwonian_{t}^{i}\right\}_{1\leq i\leq M}\) samples.}
        \For{$t = (n-1)$ \KwTo $1$}{
        \(\alpha_{t}^{M}=\argmin_\alpha \frac{1}{M} \sum_{m=1}^M |y_{t+1}^{n,M}(\brwonian_{t+1}^{m})  \frac{\Delta \brwonian_t^{m}}{\Delta_t} - \alpha \cdot \Phi_{t}(\brwonian_{t}^{m}) |^2\)
        
        \(z_{t}^{n,M}(\cdot)= \alpha_{t}^{M} \cdot \Phi_{t}(\cdot) \)
        
        \(\alpha_{0,t}^{M} = \argmin_\alpha \frac{1}{M} \sum_{m=1}^M |y_{t+1}^{n,M}(\brwonian_{t+1}^{m}) 
        + \Delta_t \hat{f}(t_t,\scorenet(y_{t+1}^{n,M}(\brwonian_{t+1}^{m})),z_{t}^{n,M}(\brwonian_{t}^{m})) -  \alpha \cdot \Phi_{0,t}(\brwonian_{t}^{m}) |^2\);
        \(y_{t}^{n,M}(\cdot)= \alpha_t^{M} \cdot \Phi_{0,t}(\cdot)\)
         }
        \(Y_0^{\pi,M}=\frac{1}{M} \sum_{m=1}^M (y_{1}^{n,M}(\brwonian_{1}^{m}) + \Delta_1 \hat{f}(t_0,\scorenet(y_{1}^{n,M}(\brwonian_{1}^{m})),z_{1}^{n,M}(\brwonian_{1}^{m}))) \)
         \caption{The Regression Monte Carlo-based BSDE Diffusion Solver.}
        \end{algorithm}
\end{figure}

\begin{proposition}
\label{corollary;flips}
The generator function of BSDE-based Diffusion $\hat{f}(\scorenet(Y_t, t), Z_t)$ is Lipschitz continuous.
\end{proposition}
Another necessary condition for the well-posedness of the BSDE-based diffusion model is the quadratically integrability of the composition function:
\begin{proposition}
\label{corollary;integrable}
The composition function $\hat{f}(\scorenet(Y_t, t), Z_t)$ is a square integrable stochastic process in $[0, T]$. 
\end{proposition}
The two propositions guarantee the well-posedness of the BSDE-based diffusion process constructed by the score network, which leading the following main theorem:
\begin{theorem}
\label{theorem;wellposed}
Under the assumption that the propositions \ref{corollary;integrable} and \ref{corollary;flips} hold true for the BSDE-based diffusion model defined in \ref{def;bsdescore}, the solution to the BSDE, guided by the Lipschitz score network $\scorenet$, exists and is uniquely defined by the stochastic processes $Y_t$ and $Z_t$.
\end{theorem}
The aforementioned theorem, \ref{theorem;wellposed}, can be demonstrated by using Pardoux-Peng's theorem of BSDEs \cite{PARDOUX199055}.

\subsection{Sampling} \label{forwardgeneration}
Given a BSDE in Eq. \ref{def;bsdescore}, according to Theorem \ref{theorem;wellposed}, we can determine two unique stochastic processes $Y_t$ and $Z_t$ to generate a target data point $\xi$. Conversely, by manipulating these two processes, we can control and analyze the score model for practical applications. There are three ways to operate the generative process: perturbing the initial state $Y_0$, changing the measure for $Z_t$ (via Girsanov transformation), or implementing both strategies. {The detailed algorithms are given in the Appendix.}

\subsubsection{$Y_0$-neighborhood Sampling:} Given the initial point $Y_0$ obtained by solving the BSDEs, we can sample randomly in the neighborhood of $Y_0$ to alter the stochastic path of $Y_t$. By integrating the forward SDE over time, we can generate similar data at the final state $\hat{Y_T} \sim \xi$. 

\subsubsection{$Z$-Girsanov Transformation:} Another approach for controlling the generation process involves altering the probability measure of the original diffusion process, which can be achieved by manipulating the distribution of $Z$. The Girsanov theorem governs the relationship between the original diffusion and the transformed diffusion. 
\section{Solving The BSDE-based Diffusion Model}
In practice, the vast majority of BSDEs we encounter do not possess analytical solutions. This is particularly true for our BSDE-based Diffusion, which is a non-linear BSDEs that include neural network terms. As a result, numerical solution methods for BSDEs are crucial for solving BSDE-based diffusion models. 
Common numerical algorithms include the approximate discrete martingale method \cite{nakayama20022, BOUCHARD2004175}, Picard iteration method \cite{peng1999linear, martinez2011numerical}, Reference ODEs-based method \cite{refode}, and Feynman-Kac PDE approximation method \cite{feynman, ma2002representation}, Least-squares regression-based methods \cite{bsde_discrete, Lin_regress}, Crank-Nicolson scheme \cite{Wang2009}, among others. In recent years, with the rapid development of deep learning in the field of numerical computation, methods for solving non-linear BSDEs using neural networks have been proposed \cite{naito2020acceleration, deepbsde}. Here, we present one representative method from each of the non-deep learning-based BSDE solver and deep learning-based approaches. A detailed survey about BSDE solvers was presented by Chessari \etal~\cite{Chessari_2023}.
\subsection{BSDEs Solver}
\begin{figure}
  \centering
  \vspace{-4mm}
  \label{fig;lips}
      \caption{10-folds Comparison Between the Networks.}
  \includegraphics[width=\textwidth]{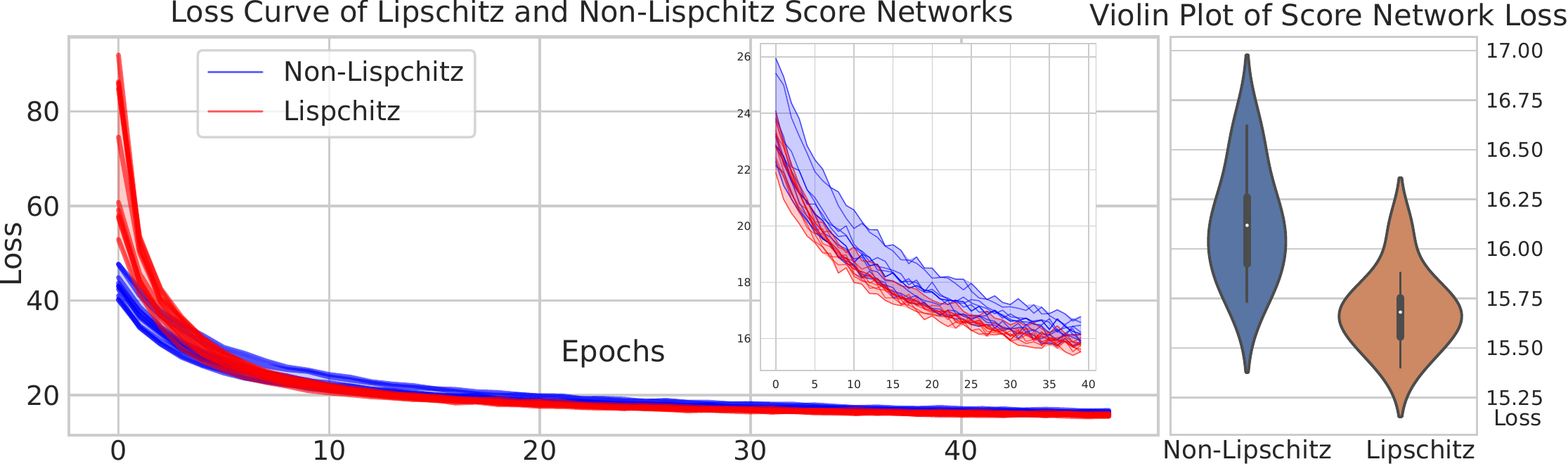}
\end{figure}
We can use the representative least-squares regression-based methods \cite{bsde_discrete, Lin_regress, Chessari_2023, longstaff2001valuing} for solving the proposed BSDE-based diffusion model. We use a discretization time of $\Delta_t=T/n$, where $t\in {0,1,\cdots,n}$, such that the incremental size of Brownian motion is given by $\Delta \brwonian_t$. To approximate the solution of the BSDE, a polynomial $\Phi$ is utilized to project the conditional expectations onto a set of basis polynomials. The CE is computed using a MCMC simulation of $M$ random paths in algorithm \ref{numsolver}. The final estimate for the initial data point $Y_0$ is obtained by averaging the updated values of the process $Y$ at $t=1$. 
We can also use deep learning-based numerical solvers.
A representative method is known as \textit{DeepBSDE} \cite{deepbsde, deepbsde2}, which offers significant flexibility by utilizing the gradient of the control process $Z$. 

The \textit{DeepBSDE} learns $Z$ through a multi-layer neural network (NN) $\Phi(t,x;\beta)$,
where $\beta$ represents the NN parameters. 
\section{Experiments and Applications}

\subsection{Empirical Analysis of Lipschitz Network for Score Matching}\label{exp;lipsnet}
We propose building the Lipschitz network $S_\theta$ based on a U-net architecture, which shows promising performance for score matching. 
Our objective is to bound the Lipschitz constant of the Noise Conditional Score U-net \cite{songNoise} to jointly estimate the scores of data distributions while also ensuring the convergence condition for the Diffusion through BSDEs.
To achieve this, we apply spectral normalization \cite{miyato2018spectral} and noise conditioning \cite{songNoise} techniques on a conventional U-net.\cite{unet} Fig. \ref{fig:comparison} presents a comparison of the score matching error for Lipschitz and Non-Lipschitz models on MNIST and CIFAR datasets. For each model and dataset, the table shows the mean and variance of the MSE of running the models 10 times, with a Mann–Whitney U significance test available. The test shows the performance improvement is statistically significant with both \textit{p-value<0.01}. The results showed that the Lipschitz network had a lower MSE compared to the Non-Lipschitz network on both datasets, with a statistically significant difference. Specifically, the Lipschitz network had an average MSE of 15.69 (standard deviation of 0.19) on MNIST and 219.20 (standard deviation of 1.31) on CIFAR, while the Non-Lipschitz network had an average MSE of 16.12 (standard deviation of 0.27) on MNIST and 221.31 (standard deviation of 0.90) on CIFAR. The Mann-Whitney U test showed a U value of 91.0 and a p-value of 0.0022, indicating a significant difference between the two networks' performance. These results suggest that the Lipschitz network performs better in terms of MSE on these datasets compared to the Non-Lipschitz network.
Qualitatively, we see from Fig. \ref{fig;lips} the Lipschitz model performs better on both datasets compared to the Non-Lipschitz model.

\subsection{BSDEs-based Diffusion for Conditioning Generation}
For illustration, we utilize a pre-trained Variance Exploding (VE) Stochastic Differential Equation (SDE) \cite{song2021scorebased}, given by $d X_t = \sigma_t d\brwonian_t, t\in[0,1]$, to train a score model on the MNIST dataset. 
\begin{figure}
    \centering
    \includegraphics[width=0.8\textwidth]{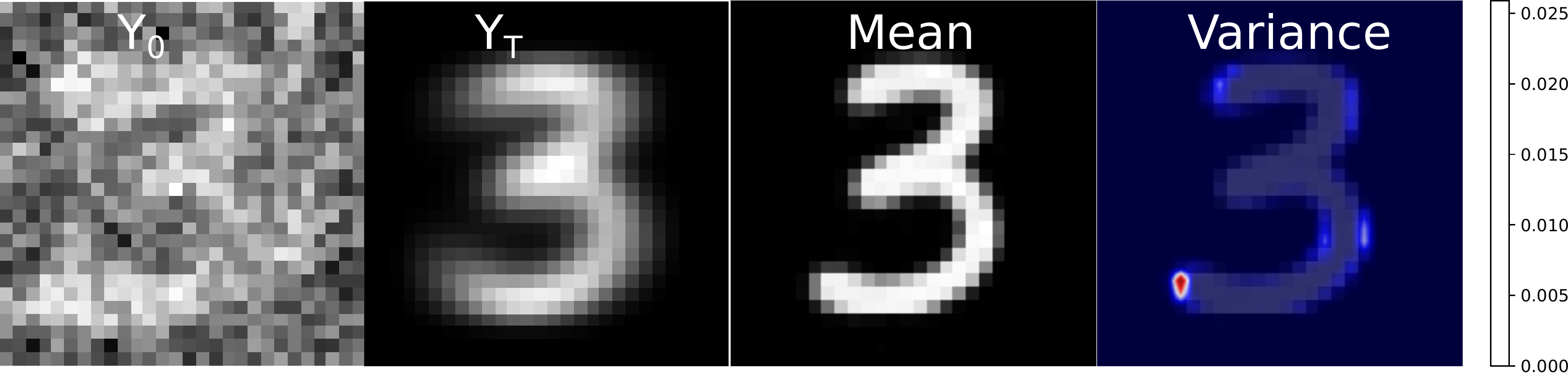}
    \caption{BSDE-based Diffusion for Uncertainty Quantification}
    \label{fig;uq}
\end{figure}
We subsequently employ a simple linear function as our generator function. Recall that the $p(Y_1=\xi) \sim p(X_1)$ (see Def. \ref{def;bsdescore}), by solving the BSDE and utilizing the sampling techniques of \textit{$Y_0$-neighborhood Sampling} and \textit{$Z$-Girsanov Transformation} as presented in Section \ref{forwardgeneration}, we can derive various applications of BSDE-based diffusion.
\begin{figure}[!ht]
    \centering
    \includegraphics[width=\textwidth]{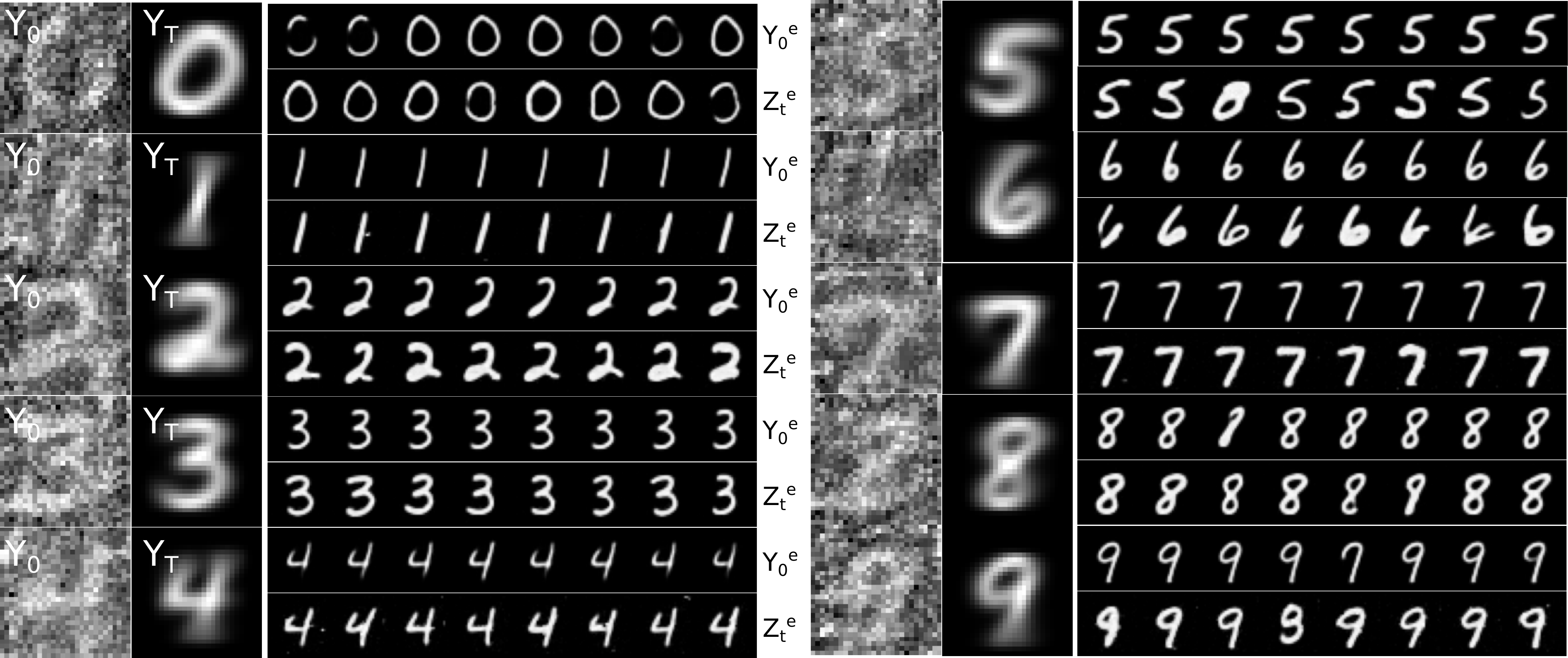}
    \vspace{-3mm}
    \caption{Demonstration of the BSDE-based Diffusion for various tasks on the MNIST dataset.}
    \vspace{-3mm}
    \label{fig:application}
\end{figure}
\textbf{Inversion:} Analogous to the GAN inversion technique that aims to map a given data point back to the latent space of a pre-trained generator \cite{xia2022gan}, the BSDE-based diffusion model can be utilized to invert a data point $Y_{T=1}$ (i.e., an image $Y_{T}$ of digit '6' from the MNIST) back to an initial data point $Y_0$, which represents a unique encoding for generating the target image $Y_{T}$ of digit '6', as guaranteed by Theorem \ref{theorem;wellposed}. To achieve this, we solve the BSDE using a numerical BSDE solver, which yields the stochastic processes $(Y, Z)$ that converge to $Y_1=\xi$.

\textbf{Unsupervised or Semi-supervised Controllable Generation:} Depending on the availability of prompt data points $Y_{T}$, controllable diffusion using BSDE can be either unsupervised or semi-supervised. In the former case, we can sample the forward process to generate guidance and utilize it to produce the desired class of data points $\hat{Y}_T$. As illustrated in Fig. \ref{fig:application} (left panel), to control the generation, we can use (1) \textit{$Y_0$-neighborhood Sampling} by perturbing the initial noise encoding $Y_0 + \lambda_Y \epsilon; \epsilon \sim \mathcal{N}(0, 1), \lambda_Y \geq 0$, (2) \textit{$Z$-Girsanov Transformation} by amplifying a gain factor $\lambda_Z \geq 1$ to modify the stochastic process $Z$ of the original BSDE $\hat{Z_t} = \lambda_Z Z_t | Y_1 = \xi$, or (3) use (1) and (2) jointly.

\textbf{Time-independent Guided Generation:} The use of guidance in diffusion-based image synthesis reduces the need for manual input, resulting in more realistic synthesized outcomes. With our proposed approach, there is no need to specify the initial time for diffusion, which is a hyperparameter of conventional SDE-based guided diffusion approaches \cite{meng2022sdedit}. This time-independent method is highly desirable, as it eliminates the need for tedious manual adjustments and allows for more efficient and effective image synthesis.

\textbf{BSDE-based Diffusion for Uncertainty Quantification}
The utilization of the BSDE-based diffusion model facilitates uncertainty quantification (UQ) in several generation-based tasks. Our BSDE-based diffusion enables UQ with as few as one sample by conducting MCMC for the $Z$ process, which evaluates the uncertainty in the reconstruction (Fig. \ref{fig;uq}).
\section{Discussion and limitations}
In this paper, we have presented a novel approach for modeling diffusion processes using Backward Stochastic Differential Equations (BSDEs) and Lipschitz networks for score matching. We have demonstrated its potential for solving practical problems, such as inversion processes in score-based learning, and established theoretical guarantees for the proposed BSDE-based Diffusion framework. However, the speed and accuracy of the BSDE solver is a significant limitation of the proposed framework. Future work should explore more computationally efficient solvers and adaptive methods to determine the gain coefficients for better separating the marginal distribution space. Moreover, the specific impact of the Lipschitz continuity of the score network should be studied in more depth.

\bibliographystyle{plain}
\bibliography{ref}


\end{document}